\documentclass{article}

\usepackage[preprint,nonatbib]{neurips_2022}

\usepackage[utf8]{inputenc} 
\usepackage[T1]{fontenc}    
\usepackage{hyperref}       
\usepackage{url}            
\usepackage{booktabs}       
\usepackage{amsfonts}       
\usepackage{nicefrac}       
\usepackage{microtype}      
\usepackage{xcolor, colortbl}         
\usepackage{etoolbox}
\usepackage{xspace}
\usepackage{graphicx}
\usepackage{subcaption}
\usepackage{siunitx}
\usepackage{multirow}
\usepackage{enumitem}
\setlist[itemize]{leftmargin=1em}

\newcommand{\lang}{\textsc{Nomos}\xspace}
\newcommand{\credit}{GermanCredit\xspace}
\newcommand{\lunar}{LunarLander\xspace}
\newcommand{\compas}{COMPAS\xspace}
\newcommand{\hotel}{HotelReview\xspace}
\newcommand{\mnist}{MNIST\xspace}
\newcommand{\speech}{SpeechCommand\xspace}
\newcommand{\pifuzz}{$\pi$-fuzz\xspace}

\newcommand{\eg}{\textit{e.g.}\xspace}
\newcommand{\ie}{\textit{i.e.}\xspace}

\newcommand{\h}[1]{{\textbf{#1}}}
\newcommand{\figref}[1]{Fig.~\ref{#1}}
\definecolor{gray}{rgb}{0.5, 0.5, 0.5}
\definecolor{light-gray}{gray}{0.77}
\definecolor{BrickRed}{rgb}{0.8, 0.25, 0.33}
\definecolor{Black}{rgb}{0.0, 0.0, 0.0}
\definecolor{DarkBlue}{rgb}{0.0, 0.0, 0.55}
\definecolor{Crimson}{rgb}{0.86, 0.08, 0.24}
\definecolor{SlateGrey}{rgb}{0.44, 0.5, 0.56}
\definecolor{lightorange}{HTML}{FFB74D}
\definecolor{blue}{rgb}{0.0, 0.0, 1.0}
\definecolor{magenta}{rgb}{0.79, 0.08, 0.48}
\definecolor{darkgreen}{rgb}{0.0, 0.2, 0.13}
\definecolor{darkraspberry}{rgb}{0.53, 0.15, 0.34}
\usepackage{listings}
\lstdefinestyle{basic}{%
  keywordstyle     = [2]\color{teal}\bfseries,%
  keywordstyle     = [3]\color{BrickRed}\bfseries,%
  keywordstyle     = [4]\color{darkraspberry}\bfseries,%
  keywordstyle     = \bfseries\color{DarkBlue},%
  commentstyle     = \ttfamily\color{Black!50}\lst@ifdisplaystyle\footnotesize\fi,%
  basicstyle       = \small\ttfamily\lst@ifdisplaystyle\footnotesize\fi,%
  emph             = {int,char,double,float,unsigned,void,bool},%
  emphstyle        = {\color{teal}\bfseries},%
  columns          = [c]fixed,%
  aboveskip        = 0mm,%
  belowskip        = 2mm,%
  keepspaces       = true,%
  mathescape       = true,%
  escapechar       = @,%
  tabsize          = 2,%
  numbers          = left,%
  numberstyle      = \tiny\color{Black!70},%
  numbersep        = 4pt,%
  stepnumber       = 1,%
  firstnumber      = 1,%
  showstringspaces = false,%
  captionpos       = b,%
  extendedchars    = true,%
  upquote          = true,%
  abovecaptionskip = 0mm,%
  belowcaptionskip = 0mm,%
  moredelim        = **[is][{\btHL[fill=light-gray]}]{Â°}{Â°},%
}

\lstdefinestyle{nomos}{%
  language         = Python,%
  style            = basic,%
  morekeywords     = [1]{var,input,output,ensures,import,requires},%
  morekeywords     = [2]{randInt,getFeat,label,strConcat,setFeat,blur,wNoise,relax,unrelax},%
  morekeywords     = [3]{MAX_INT},%
  morekeywords     = [4]{play,predict},%
  xleftmargin=1.2em,%
}

\newcommand\code[1]{\lstinline[style=nomos]{#1}}

\lstdefinestyle{python}{%
  language         = Python,%
  style            = basic,%
  xleftmargin=1.2em,%
}

\lstdefinestyle{basic-grammar}{%
  morekeywords     = [1]{<spec>,<import>,<input>,<var_decl>,<precond>,<output>,<postcond>,<code>,<scalar_expr>,<bool_expr>,<record_expr>,<bool_literal>,<int_expr>,<string_expr>,<var_name>,<model_name>,<int_literal>,<string_literal>},%
  morekeywords     = [2]{::=},%
  keywordstyle     = [2]\color{darkgreen}\bfseries,%
  morekeywords     = [3]{},%
  keywordstyle     = [3]\color{BrickRed}\bfseries,%
  morekeywords     = [4]{},%
  keywordstyle     = [4]\color{darkraspberry}\bfseries,%
  keywordstyle     = \bfseries\color{DarkBlue},%
  commentstyle     = \ttfamily\color{Black!50}\lst@ifdisplaystyle\footnotesize\fi,%
  basicstyle       = {\small\ttfamily\lst@ifdisplaystyle\scriptsize\fi},%
  emph             = {},%
  emphstyle        = {\color{teal}\bfseries},%
  columns          = [c]fixed,%
  literate=%
    {|}{$\mid$}1%
    {::=}{$::$$=$}3,%
  aboveskip        = 0mm,%
  belowskip        = 2mm,%
  keepspaces       = true,%
  mathescape       = true,%
  escapechar       = @,%
  tabsize          = 2,%
  numbers          = left,%
  numberstyle      = \tiny\color{Black!70},%
  numbersep        = 4pt,%
  stepnumber       = 1,%
  firstnumber      = 1,%
  showstringspaces = false,%
  stringstyle      = \color{Crimson},%
  captionpos       = b,%
  extendedchars    = true,%
  upquote          = true,%
  abovecaptionskip = 0mm,%
  belowcaptionskip = 0mm,%
  alsoletter       = {<>:=|{}}%
  moredelim        = **[is][{\btHL[fill=light-gray]}]{Â°}{Â°},%
}

\lstdefinestyle{ebnf}{%
  language         = Python,%
  style            = basic-grammar,%
  xleftmargin=1.2em,%
}

\newcommand\ebnf[1]{\lstinline[style=ebnf]{#1}}

\title{Specifying and Testing $k$-Safety Properties\\for Machine-Learning Models}

\author{%
  Maria Christakis \\
  MPI-SWS \\
  Germany \\
  \texttt{maria@mpi-sws.org} \\
  \And
  Hasan Ferit Eniser \\
  MPI-SWS \\
  Germany \\
  \texttt{hfeniser@mpi-sws.org} \\
  \AND
  J\"org Hoffmann \\
  Saarland University, Saarland Informatics Campus \\
  German Research Center for Artificial Intelligence (DFKI) \\
  Germany \\
  \texttt{hoffmann@cs.uni-saarland.de} \\
  \And
  Adish Singla \\
  MPI-SWS \\
  Germany \\
  \texttt{adishs@mpi-sws.org} \\
  \And
  Valentin W\"ustholz \\
  ConsenSys \\
  Germany \\
  \texttt{valentin.wustholz@consensys.net} \\
}
\begin{document}
\maketitle

\newtoggle{longversion}
\settoggle{longversion}{true}

\begin{abstract}
Machine-learning models are becoming increasingly prevalent in our lives, for instance assisting in image-classification or decision-making tasks. Consequently, the reliability of these models is of critical importance and has resulted in the development of numerous approaches for validating and verifying their robustness and fairness. However, beyond such specific properties, it is challenging to specify, let alone check, general functional-correctness expectations from models. In this paper, we take inspiration from specifications used in formal methods, expressing functional-correctness properties by reasoning about $k$ different executions---so-called $k$-safety properties. Considering a credit-screening model of a bank, the expected property that "if a person is denied a loan and their income decreases, they should still be denied the loan" is a 2-safety property. Here, we show the wide applicability of $k$-safety properties for machine-learning models and present the first specification language for expressing them. We also operationalize the language in a framework for automatically validating such properties using metamorphic testing. Our experiments show that our framework is effective in identifying property violations, and that detected bugs could be used to train better models.
\end{abstract}
\section{Introduction}\label{sec.intro}

Due to the impressive advances in machine learning and the unlimited availability of data, machine-learning (ML) models, \eg, neural networks, are rapidly becoming prevalent in our lives, for instance by assisting in image-classification or decision-making tasks. As a result, there is growing concern about the reliability of these models in performing such tasks. For example, it could be disastrous if an autonomous vehicle misclassifies a street sign, or if a recidivism-risk algorithm, which predicts whether a criminal is likely to re-offend, is unfair with respect to race. The research community is, of course, aware of these issues and has devised numerous techniques to validate and verify robustness and fairness properties of machine-learning models (\eg, \cite{HuangKwiatkowska2017,GehrMirman2018,SinghGehr2019-Domain,AlbarghouthiDAntoni2017-Verification,BastaniZhang2019,UrbanChristakis2020,CarliniWagner2017-Robustness,GoodfellowShlens2015,MadryMakelov2018,GalhotraBrun2017,UdeshiArora2018,TramerAtlidakis2017}).

Beyond such specific properties however, it is challenging to express general functional-correctness expectations from such models, let alone check them, \eg, how can we specify that an image classifier should label images correctly? We take inspiration from specifications used in formal methods---so-called \emph{hyperproperties}~\cite{ClarksonSchneider2008}---capturing functional-correctness properties by simultaneously reasoning about multiple system executions. For example, consider a credit-screening model of a bank. The expected property that "if a person is denied a loan and their income decreases, they should still be denied the loan", or conversely "if a person is granted a loan and their income increases, they should still be granted the loan", is a \emph{2-safety} hyperproperty---we need two model executions to validate its correctness. In contrast, the property that "a person with no income should be denied a loan" is a standard (1-)safety property since it can be validated by individual model executions. Overall, \emph{$k$-safety hyperproperties} generalize standard safety properties in that they require reasoning about $k$ different executions.

\textbf{Examples.}
To demonstrate the wide applicability of $k$-safety properties for ML models, we use examples from five distinct domains throughout this paper:
\begin{itemize}
\item \emph{Tabular data.} Consider the \compas dataset~\cite{COMPAS}, which determines how likely criminals are to re-offend. An expected hyperproperty for models trained on \compas could be that "if the number of committed felonies for a given criminal increases, then their recidivism risk should not decrease". Note that this is essentially monotonicity in an input feature, a special case of the hyperproperties we consider here.

\item \emph{Images.} Using the \mnist dataset~\cite{MNIST}, which classifies images of handwritten digits, an expected hyperproperty could be that "if a blurred image is correctly classified, then its unblurred version should also be correctly classified". Note that this is \emph{not} monotonicity in a feature as whether or not an image is blurred does not constitute part of the model input (\ie, the image); instead, blurring may affect most, if not all, pixels.

\item \emph{Speech.} Similarly, for the \speech dataset~\cite{Warden2018}, which classifies short spoken commands, an expected hyperproperty could be that "if a speech command with white noise is correctly classified, then its non-noisy version should also be correctly classified".

\item \emph{Natural language.} The \hotel dataset~\cite{HotelReview} is used for sentiment analysis of hotel reviews. An expected hyperproperty could be that "if a review becomes more negative, then the sentiment should not become more positive". Note that, again, making a review more negative may significantly affect the model input.

\item \emph{Action policies.} \lunar is a popular Gym~\cite{BrockmanCheung2016} environment consisting of a 2D-world with an uneven lunar surface and a reinforcement-learning (RL) lander, which initially appears far above the surface and moves downward. The goal is to navigate and land the lander on its two legs; if the body ever touches the surface, the lander crashes. An expected hyperproperty could be that "if the lander lands successfully, then decreasing the surface height (thus giving the lander more time to land) should also result in landing successfully". Here, even a seemingly simple change to the initial game state may result in significant changes to subsequent states since the policy is invoked repeatedly during the game.
\end{itemize}
Note that, in practice, such properties are defined by users, thus expressing model expectations that are deemed important in their particular usage scenario.

\textbf{Related work.}
There is work on expressing $k$-safety properties for programs~\cite{SousaDillig2016}, but no prior work has explored how to specify such properties for ML models and how to leverage these specifications for automated testing.
Numerous techniques verify specific functional-correctness properties of models, such as robustness (\eg, \cite{HuangKwiatkowska2017,GehrMirman2018,SinghGehr2019-Domain,BerradaDathathri2021,WangZhang2021}), fairness (\eg, \cite{AlbarghouthiDAntoni2017-Verification,BastaniZhang2019,UrbanChristakis2020}), and others (\eg, \cite{KatzBarrett2017,WangPei2018-SecurityAnalysis}). There are also approaches for validating such properties (\eg, \cite{CarliniWagner2017-Robustness,GoodfellowShlens2015,MadryMakelov2018,GalhotraBrun2017,UdeshiArora2018,TramerAtlidakis2017}). Although certain popular robustness and fairness properties do, in fact, constitute 2-safety properties (\eg, slightly perturbing the pixels of an image should not change its classification, or changing the race of a criminal should not make them more or less likely to re-offend), none of this work targets general hyperproperties.
The most relevant work is by Sharma and Wehrheim~\cite{SharmaWehrheim2020}, who introduce verification-based testing of monotonicity in ML models. As indicated above, a model is said to be monotone with respect to an input feature if an increase in the feature implies an increase in the model's prediction, \eg, the higher the income, the larger the loan.
In addition, Deng et al.~\cite{DengZheng2020,DengLou2021} propose an approach for testing image-based autonomous-vehicle models against safety properties defined in domain-specific behaviour templates that resemble natural language.

\textbf{Approach and contributions.}
In this paper, we show the wide applicability of $k$-safety properties for ML models. We design a declarative, domain-agnostic specification language, \lang ("law" in Greek), for writing them. In contrast to existing approaches, \lang can express general $k$-safety properties capturing \emph{arbitrary} relations between more than one input-output pair; these subsume the more specific relations of robustness, fairness, and monotonicity.
Going a step further, we design a fully automated framework for validating \lang properties using \emph{metamorphic testing}~\cite{ChenCheung1998,SeguraFraser2016}. On a high-level, our framework takes as input the model under test and a set of $k$-safety properties for the model. As output, it produces test cases for which the model violates the specified properties. Note that a single test case for a $k$-safety property consists of $k$ concrete inputs to the model under test. Under the hood, the \emph{harness generator} component of the framework compiles the provided \lang properties into a \emph{test harness}, \ie, software that tests the given model against the properties. The harness employs a \emph{test generator} and an \emph{oracle} component, for generating inputs to the model using metamorphic testing and for detecting property violations, respectively.

In summary, this paper makes the following key contributions:
\begin{itemize}
    \item We present \lang, the first specification language for expressing general $k$-safety hyperproperties for ML models.
    \item We demonstrate the wide applicability of such properties through case studies from several domains and the expressiveness of our language in capturing them.    
    \item We design and implement a fully automated framework for validating such properties using metamorphic testing.
    \item We evaluate the effectiveness of our testing framework in detecting property violations across a broad range of different domains. We also perform a feasibility study to showcase how such violations can be used to improve model training.
\end{itemize}

\section{\lang Specification Language}\label{sec.lang}

\lang allows a user to specify $k$-safety properties over source code invoking an ML model under test. On a high level, a \lang specification consists of three parts: (1)~the \emph{precondition}, (2)~the source code---Python in our implementation---invoking the model, and (3)~the \emph{postcondition}. Pre- and postconditions are commonly used in formal methods, for instance, in Hoare logic~\cite{Hoare1969} and design by contract~\cite{Eiffel}. Here, we adapt them for reasoning about ML models and $k$-safety properties.

The precondition captures the conditions under which the model should be invoked, allowing the user to express arbitrary relations between more than one model input. It is expressed using zero or more \code{requires} statements, each capturing a condition over inputs; the logical conjunction of these conditions constitutes the precondition. The source code may be arbitrary code invoking the model one or more times to capture $\mathit{k}$ input-output pairs. Finally, the postcondition captures the safety property that the model is expected to satisfy. It is expressed using zero or more \code{ensures} statements, each taking a condition that, unlike for the precondition, may refer to model outputs; the logical conjunction of these conditions constitutes the postcondition.

\begin{figure*}[t!]
\begin{subfigure}[b]{.54\textwidth}
\begin{lstlisting}[style=nomos]
input x1; @ \label{line:compas-input} @
var v1 := getFeat(x1, 1); @ \label{line:compas-var-begin} @
var v2 := v1 + randInt(1, 10);
var x2 := setFeat(x1, 1, v2); @ \label{line:compas-var-end} @
requires v2 <= 20; @ \label{line:compas-prec} @
output d1; @ \label{line:compas-out1} @
output d2; @ \label{line:compas-out2} @
{ @ \label{line:compas-code-begin} @
  d1 = predict(x1)
  d2 = predict(x2)
} @ \label{line:compas-code-end} @
# 0-low, 1-medium, 2-high risk
ensures d1 <= d2; @ \label{line:compas-post} @
\end{lstlisting}
\caption{\compas 2-safety hyperproperty.}
\label{fig:compas-example}
\end{subfigure}
\begin{subfigure}[b]{.45\textwidth}
\begin{lstlisting}[style=nomos]
input x1; @ \label{line:mnist-input} @
var x2 := blur(x1); @ \label{line:mnist-blur} @
var v1 := label(x1); @ \label{line:mnist-label} @
output d1;
output d2;
{
  d1 = predict(x1)
  d2 = predict(x2)
}
ensures d2==v1 ==> d1==v1; @ \label{line:mnist-post} @
\end{lstlisting}
\caption{\mnist 2-safety hyperproperty.}
\label{fig:mnist-example}
\end{subfigure}\\\\
\begin{subfigure}[b]{.54\textwidth}
\begin{lstlisting}[style=nomos]
input x1; @ \label{line:hotel-input1} @ 
input x2; @ \label{line:hotel-input2} @
var v1 := getFeat(x1, 1); @ \label{line:hotel-neg1} @
var v2 := getFeat(x2, 1); @ \label{line:hotel-neg2} @
var v3 := strConcat(v1, v2); @ \label{line:hotel-concat1} @
var x3 := setFeat(x1, 1, v3); @ \label{line:hotel-concat2} @
output d1;
output d3;
{
  d1 = predict(x1)
  d3 = predict(x3)
}
# 0-pos, 1-neg
ensures d1 <= d3; @ \label{line:hotel-post} @
\end{lstlisting}
\caption{\hotel 2-safety hyperproperty.}
\label{fig:hotel-example}
\end{subfigure}%
\begin{subfigure}[b]{.45\textwidth}
\begin{lstlisting}[style=nomos]
input s1; @ \label{line:lunar-input} @
var s2 := relax(s1); @ \label{line:lunar-var} @
output o1; @ \label{line:lunar-out1} @
output o2; @ \label{line:lunar-out2} @
{ @ \label{line:lunar-code-begin} @
  o1, o2 = 0, 0
  for _ in range(10):
    rs = randInt(0, MAX_INT) @ \label{line:lunar-new-rs} @
    o1 += play(s1, rs) @ \label{line:lunar-rs1} @
    o2 += play(s2, rs) @ \label{line:lunar-rs2} @
} @ \label{line:lunar-code-end} @
# 0-lose, 1-win
ensures o1 <= o2; @ \label{line:lunar-post} @
\end{lstlisting}
\caption{\lunar 20-safety hyperproperty.}
\label{fig:lunar-example}
\end{subfigure}
\caption{Example $k$-safety specifications in \lang.}
\label{fig:examples}
\end{figure*}

\textbf{Examples.}
As an example, consider the \lang specification of \figref{fig:compas-example} expressing the \compas property described earlier.
On line~\ref{line:compas-input}, we specify that we need an input \code{x1}, \ie, a criminal. Lines~\ref{line:compas-var-begin}--\ref{line:compas-var-end} get the first feature of \code{x1}, which corresponds to the number of felonies, and assign it to variable \code{v1}; in variable \code{v2}, we increase this number, and create a new criminal \code{x2} that differs from \code{x1} only with respect to this feature, \ie, \code{x2} has committed more felonies than \code{x1}. Line~\ref{line:compas-prec} specifies a precondition that the new criminal's felonies should not exceed a sensible limit. Lines~\ref{line:compas-out1}--\ref{line:compas-out2} declare two outputs, \code{d1} and \code{d2}, that are assigned the model's prediction when calling it with criminal \code{x1} and \code{x2}, respectively (see block of Python code on lines~\ref{line:compas-code-begin}--\ref{line:compas-code-end}). Finally, on line~\ref{line:compas-post}, we specify the postcondition that the recidivism risk of criminal \code{x2} should not be lower than that of \code{x1}.

\figref{fig:mnist-example} shows the \mnist specification. Given an image \code{x1} (line~\ref{line:mnist-input}), image \code{x2} is its blurred version (line~\ref{line:mnist-blur}), and variable \code{v1} contains its correct label (line~\ref{line:mnist-label}), \eg, retrieved from the dataset. Note that functions such as \code{blur} and \code{label} extend the core \lang language and may be easily added by the user.
The postcondition on line~\ref{line:mnist-post} says that if the blurred image is correctly classified, then so should the original image. Note that we defined a very similar specification for the \speech property---instead of \code{blur}, we used function \code{wNoise} adding white noise to audio.

The \hotel specification is shown in \figref{fig:hotel-example}. Note that a hotel review consists of a positive and a negative section, where a guest describes what they liked and did not like about the hotel, respectively. On lines~\ref{line:hotel-input1}--\ref{line:hotel-input2}, we obtain two reviews, \code{x1} and \code{x2}, and in variables \code{v1} and \code{v2} on lines~\ref{line:hotel-neg1}--\ref{line:hotel-neg2}, we store their negative sections (feature \code{1} retrieved with function \code{getFeat}). We then create a third review, \code{x3}, which is the same as \code{x1} except that its negative section is the concatenation of \code{v1} and \code{v2} (lines~\ref{line:hotel-concat1}--\ref{line:hotel-concat2}). The postcondition on line~\ref{line:hotel-post} checks that the detected sentiment is not more positive for review \code{x3} than for \code{x1}.

Finally, consider the \lunar specification in \figref{fig:lunar-example}. On line~\ref{line:lunar-input}, we obtain an input \code{s1}, which is an initial state of the game. Line~\ref{line:lunar-var} "relaxes" this state to obtain a new state \code{s2}, which differs from \code{s1} only in that the height of the lunar surface is lower.
In the block of Python code that follows (lines~\ref{line:lunar-code-begin}--\ref{line:lunar-code-end}), we initialize outputs \code{o1} and \code{o2} to zero and play the game from each initial state, \code{s1} and \code{s2}, in a loop; \code{o1} and \code{o2} accumulate the number of wins. We use a loop because the environment is stochastic---firing an engine of the lander follows a probability distribution. Therefore, by changing the environment random seed \code{rs} on line~\ref{line:lunar-new-rs}, we take stochasticity into account. In each loop iteration however, we ensure that the game starting from \code{s2} is indeed easier, \ie, that stochasticity cannot make it harder, by using the same seed on lines~\ref{line:lunar-rs1}--\ref{line:lunar-rs2}. Note that function \code{play} invokes the policy multiple times (\ie, after every step in the game simulator).
Finally, line~\ref{line:lunar-post} ensures that, when playing the easier game (starting with \code{s2}), the number of wins should not decrease. Since this property depends on 20 model invocations, it is a 20-safety property! Conversely, we can also make the game harder by "unrelaxing" the original initial state, \ie, increasing the surface height. In such a case, when playing the harder game, the number of wins should not exceed the original number of wins.

\begin{figure}[t]
\begin{lstlisting}[style=ebnf]
<spec>        ::= { <import> } <input> { <input> } { <var_decl> } { <precond> } @ \label{line:grammar-spec-start} @
                  { <output> } "{" <code> "}" { <postcond> } @ \label{line:grammar-spec-end} @
<import>      ::= "import" <model_name> ";" @ \label{line:grammar-import} @
<input>       ::= "input" <var_name> ";" @ \label{line:grammar-input} @
<var_decl>    ::= "var" <var_name> ":=" <scalar_expr> ";"   @ \label{line:grammar-var-start} @
                | "var" <var_name> ":=" <record_expr> ";"   @ \label{line:grammar-var-end} @
<precond>     ::= "requires" <bool_expr> ";" @ \label{line:grammar-precond} @
<output>      ::= "output" <var_name> ";" @ \label{line:grammar-output} @
<postcond>    ::= "ensures" <bool_expr> ";" @ \label{line:grammar-postcond} @
<scalar_expr> ::= <bool_expr>   @ \label{line:grammar-scalar-start} @
                | <int_expr>
                | <string_expr>
                | "getFeat(" <record_expr> "," <int_expr> ")"   @ \label{line:grammar-scalar-getFeat} @
                | "label(" <record_expr> ")"
                | "randInt(" <int_expr> "," <int_expr> ")"
                | "strConcat(" <string_expr> "," <string_expr> ")"   @ \label{line:grammar-scalar-end} @
<bool_expr>   ::= <bool_literal>   @ \label{line:grammar-bool-start} @
                | <var_name>
                | "!"<bool_expr>
                | <bool_expr> "&&" <bool_expr>
                | <scalar_expr> "==" <scalar_expr>
                | <scalar_expr> "<" <scalar_expr>
                | <record_expr> "==" <record_expr>   @ \label{line:grammar-bool-end} @
<record_expr> ::= <var_name>   @ \label{line:grammar-record-start} @
                | "setFeat(" <record_expr> "," <int_expr> "," <scalar_expr> ")"  @ \label{line:grammar-record-setFeat} @
                | "blur(" <record_expr> ")"
                | "wNoise(" <record_expr> ")"
                | "relax(" <record_expr> ")"
                | "unrelax(" <record_expr> ")"   @ \label{line:grammar-record-end} @
\end{lstlisting}
  \caption{The \lang grammar.}
  \label{fig:grammar}
\end{figure}

\textbf{Grammar.}
\figref{fig:grammar} provides a formal grammar for \lang (in a variant of extended Backus-Naur form). The top-level construct is \ebnf{<spec>} on lines~\ref{line:grammar-spec-start}--\ref{line:grammar-spec-end}. It consists of zero or more \code{import} statements---the curly braces denote repetition---to import source-code files containing custom implementations for domain-specific functions, \eg, \code{blur} or \code{wNoise}, one or more input declarations, variable declarations, preconditions, output declarations, the source-code block, and postconditions.
We define these sub-constructs in subsequent rules (lines~\ref{line:grammar-input}--\ref{line:grammar-postcond}). For instance, a precondition (line~\ref{line:grammar-precond}) consists of the token \code{requires}, a Boolean expression, and a semicolon. For brevity, we omit a definition of \ebnf{<code>}; it denotes arbitrary Python code that is intended to invoke the model under test and assign values to output variables. We also omit the basic identifiers \ebnf{<model\_name>} and \ebnf{<var\_name>}.

The grammar additionally defines various types of expressions needed in the above sub-constructs. In their definitions, we use the \ebnf{|} combinator to denote alternatives. In particular, we define scalar (lines~\ref{line:grammar-scalar-start}--\ref{line:grammar-scalar-end}), Boolean (lines~\ref{line:grammar-bool-start}--\ref{line:grammar-bool-end}), and record expressions (lines~\ref{line:grammar-record-start}--\ref{line:grammar-record-end}). The latter are used to express complex object-like values, such as images or game states. In these definitions, we include extensions to the core language with domain-specific functions that support the application domains considered in this paper---\eg, \code{getFeat} and \code{setFeat} retrieve and modify record fields, respectively. Integer and string expressions are defined as expected (see appendix), and we omit the basic scalar expressions \ebnf{<bool\_literal>}, \ebnf{<int\_literal>}, and \ebnf{<string\_literal>}.

\section{Metamorphic-Testing Framework for \lang Specifications}\label{sec.tool}

\emph{Metamorphic testing}~\cite{ChenCheung1998,SeguraFraser2016} is a testing technique that addresses the lack of an existing \emph{oracle} defining correct system behavior. Specifically, given an input, metamorphic testing transforms it such that the relation between the outputs (\ie, the output of the system under test when executed on the original input and the corresponding output when executed on the transformed input) is known. If this relation between outputs does not actually hold, then a bug is detected. As a simple example, consider testing a database system; given a query as the original input, assume that the transformed input is the same query with weakened constraints. A bug is detected if the transformed query returns fewer results than the original one, which is more restrictive. So far, metamorphic testing has been used to test ML models from specific application domains, \eg, image classifiers~\cite{DwarakanathAhuja2018,TianZhong2020}, translation systems~\cite{SunZhang2022}, NLP models~\cite{MaWang2020}, object-detection systems~\cite{WangSu2020}, action policies~\cite{EniserGros2022}, and autonomous cars~\cite{TianPei2018,ZhangZhang2018}.

\begin{figure}[t]
    \centering
    \includegraphics[width=0.9\textwidth,trim={6cm 10cm 5.5cm 5cm},clip]{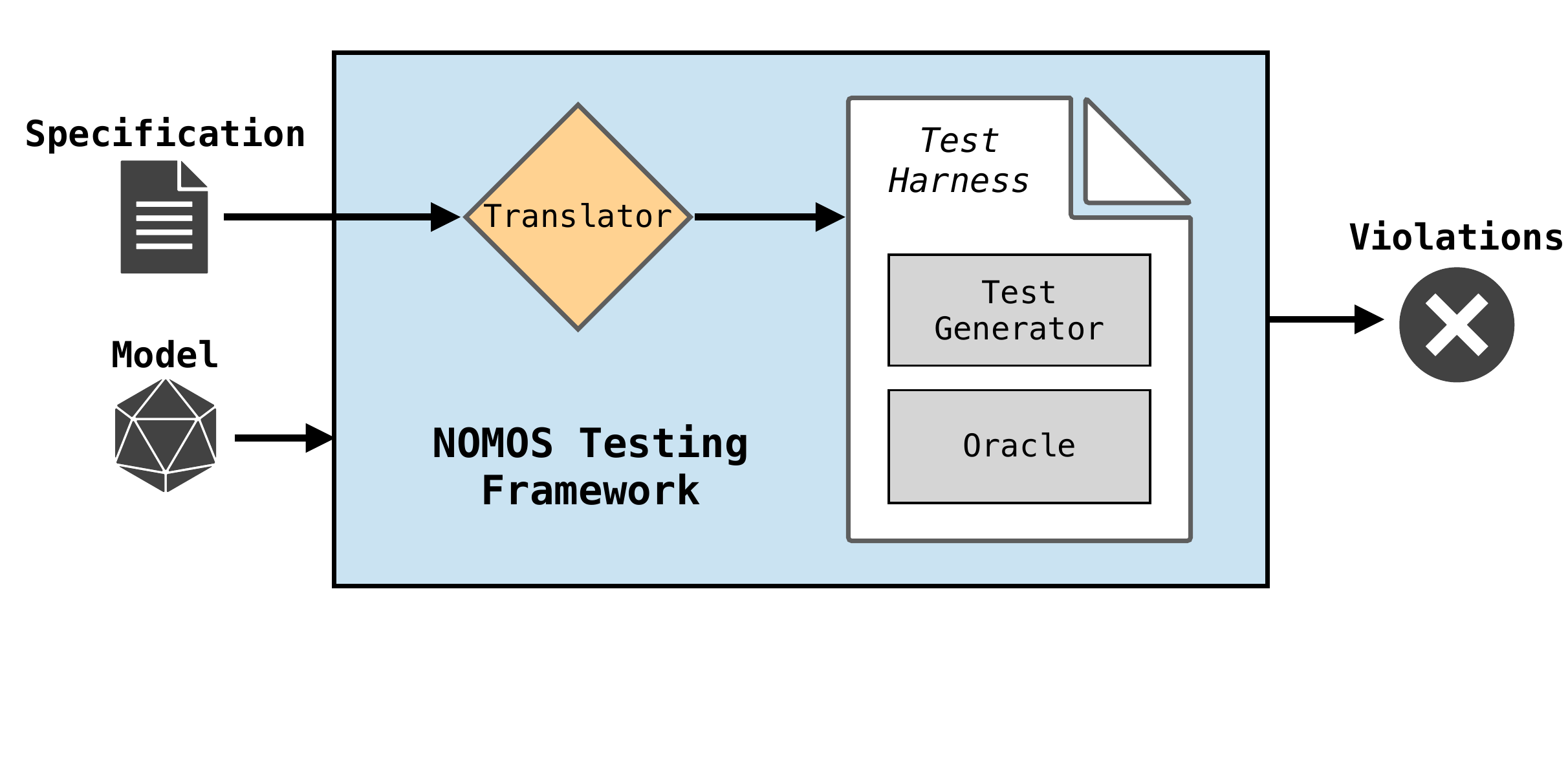}
    \vspace{-3em}
    \caption{An overview of our testing framework.}
    \label{fig:framework}
\end{figure}

In our setting, we observe that metamorphic testing is a natural choice for validating general $k$-safety properties as these also prescribe input transformations and expected output relations. For instance, in \figref{fig:compas-example}, lines~\ref{line:compas-var-begin}--\ref{line:compas-prec} describe the transformation to input \code{x1} in order to obtain \code{x2}, and line~\ref{line:compas-post} specifies the relation between the corresponding outputs. We, therefore, design the framework in \figref{fig:framework} for validating a model against a \lang specification using metamorphic testing. The output of our framework is a set of (unique) bugs, \ie, test cases revealing postcondition violations. For \figref{fig:compas-example}, a bug would comprise two concrete instances of a criminal, $c_1$ and $c_2$, such that (1)~$c_2$ differs from $c_1$ only in having more felonies, and (2)~the recidivism risk of $c_2$ is predicted to be lower than that of $c_1$.

Under the hood, the \emph{harness generator} component of the framework compiles the \lang specification into a \emph{test harness}, \ie, a Python program that tests the model against the specified properties. Our implementation parses \lang specifications using an ANTLR4~\cite{ANTLR4} grammar. After semantically checking the parsed abstract syntax tree (AST), our framework translates the AST into the Python program constituting the test harness. A snippet of the generated harness for the specification of \figref{fig:compas-example} is shown in \figref{fig:compas-harness}. The test harness employs a \emph{test generator} and an \emph{oracle} component, for generating inputs to the model using metamorphic testing and for detecting postcondition violations, respectively.

\begin{figure*}[b]
\begin{subfigure}[t]{.50\textwidth}
\begin{lstlisting}[style=python]
while budget > 0: @\label{line:budget}@

  # test generator @\label{line:input-gen-begin}@
  x1 = compas.randInput() @\label{line:rand-input}@
  v1 = compas.getFeat(x1,1) @\label{line:temp-vars-begin}@
  v2 = v1 + compas.randInt(1,10) @\label{line:rand-int}@
  x2 = compas.setFeat(x1,1,v2) @\label{line:temp-vars-end}@
  
  if not(v2 <= 20): @\label{line:prec}@
    compas.precond_violtn += 1 
    continue @\label{line:input-gen-end}@
\end{lstlisting}
\end{subfigure}%
\begin{subfigure}[t]{.50\textwidth}
\begin{lstlisting}[style=python,firstnumber=12]
  # code @\label{line:code-begin}@
  d1 = compas.predict(x1)
  d2 = compas.predict(x2) @\label{line:code-end}@

  # oracle @\label{line:oracle-begin}@
  if d1 <= d2 :
    compas.passed += 1
  else:
    compas.postcond_violtn += 1
    compas.process_bug() @\label{line:oracle-end}@
  budget -= 1
\end{lstlisting}
\end{subfigure}
\caption{Snippet of generated harness for the specification of \figref{fig:compas-example}.}
\label{fig:compas-harness}
\end{figure*}

As shown in \figref{fig:compas-harness}, the model is tested until a user-specified budget is depleted (line~\ref{line:budget}). In each iteration of this loop, the test generator creates $k$ model inputs that satisfy the given precondition, if any (lines~\ref{line:input-gen-begin}-\ref{line:input-gen-end}). Specifically, for every \code{input} declaration, the test generator randomly selects an input from a source specified in the imported files (line~\ref{line:rand-input})---note that \code{import} statements are not shown here but are defined on line~\ref{line:grammar-import} of \figref{fig:grammar}. In our evaluation, we have used both the test set and the output of an off-the-shelf fuzzer~\cite{EniserGros2022} as such input sources. The metamorphic transformation of an input can be performed through \code{var} declarations, which are compiled into temporary variables in the harness (lines~\ref{line:temp-vars-begin}--\ref{line:temp-vars-end}). Before the test generator returns the $k$ generated model inputs, the specified precondition is checked; if it is violated, the process repeats until it holds (lines~\ref{line:prec}--\ref{line:input-gen-end}).

Next, the block of Python code in the specification is executed (lines~\ref{line:code-begin}--\ref{line:code-end}), and finally the oracle component checks the postcondition (lines~\ref{line:oracle-begin}--\ref{line:oracle-end}). On line~\ref{line:oracle-end}, the oracle records each detected bug and processes it for subsequent de-duplication. In particular, for each bug, the oracle hashes any source of randomness in generating the model inputs (\ie, for the example of \figref{fig:compas-harness}, there is randomness on lines~\ref{line:rand-input} and \ref{line:rand-int}). Two bugs are considered duplicate if their hashes match, that is, if the generated model inputs are equivalent. Note that we avoid comparing model inputs directly due to their potential complexity, \eg, in the case of game states.
\section{Experimental Evaluation}\label{sec.experiments}

So far, we have demonstrated the expressiveness of \lang by specifying hyperproperties for models in diverse domains. This section focuses on evaluating the effectiveness of our testing framework in finding bugs in these models. We describe the benchmarks, experimental setup, and results. We also present a feasibility study on how detected bugs can be used to improve model training.

\textbf{Benchmarks.}
We trained models using six datasets from five application domains as follows:
\begin{itemize}
\item \emph{Tabular data.} We used the \compas~\cite{COMPAS} and \credit~\cite{GermanCredit} datasets, which we pre-process. For each dataset, we trained a fully connected neural network (NN) and a decision tree (DT). For \compas, we achieved 74\% (NN) and 72\% (DT) accuracy, and for \credit, 78\% (NN) and 70\% (DT). Note that, even though we report accuracy here, the achieved score does not necessarily affect whether a specified property holds, \ie, a perfectly accurate model could violate the property, whereas a less accurate model might not.

\item \emph{Images.} Using the \mnist dataset~\cite{MNIST}, we trained a fully connected neural network achieving 97\% accuracy.

\item \emph{Speech.} We pre-processed the \speech dataset~\cite{Warden2018} to convert waveforms to spectrograms, which show frequency changes over time. As spectrograms are typically represented as 2D-images, we trained a convolutional neural network classifying spectrogram images. The model achieves 84\% test accuracy.

\item \emph{Natural language.} For the \hotel dataset~\cite{HotelReview}, we used a pre-trained Universal Sentence Encoder (USE)~\cite{CerYang2018} to encode natural-language text into high dimensional vectors. USE compresses any textual data into a vector of size 512 while preserving the similarity between sentences. We trained a fully connected neural network of 82\% accuracy on the encoded hotel reviews. 

\item \emph{Action policies.} In \lunar~\cite{BrockmanCheung2016}, touching a leg of the lander to the surface yields reward $+100$, whereas touching the body yields $-100$; the best-case reward is over $200$.
We trained an RL policy that achieves an average reward of $175$ (after $1$ million training episodes).
\end{itemize}

\textbf{Experimental setup.}
For each of these models, we wrote one or more \lang specifications to capture potentially desired properties (see appendix for a complete list).
Each test harness used a budget of $5000$ (see line~\ref{line:budget} of \figref{fig:compas-harness}), that is, it generated $5000$ test cases satisfying the precondition, if any. We ran each harness with $10$ different random seeds to account for randomness in the testing procedure. Here, we report arithmetic means (\eg, for the number of bugs) unless stated otherwise. In all harnesses except for \lunar, the input source (\eg, line~\ref{line:rand-input} of \figref{fig:compas-harness}) is the test set. In the \lunar harness, the input source is a pool of game states that was generated by \pifuzz~\cite{EniserGros2022} after fuzzing our policy for $2$ hours.

\begin{table}[t]
\caption{Number of specified properties, violated properties, and unique bugs per dataset and model.}
\vspace{1em}
\centering
\newcolumntype{g}{>{\columncolor{light-gray!20}}r}
\newcolumntype{h}{>{\columncolor{light-gray!20}}c}
\begin{tabular}{g|h|S[table-format=1.0]|S[table-format=2.0]|S[table-format=4.1]}
\rowcolor{light-gray!50}
& & \multicolumn{2}{c|}{\textbf{Properties}} & \multicolumn{1}{c}{\textbf{Unique}}\\
\rowcolor{light-gray!50}
\multicolumn{1}{c|}{\multirow{-2}{*}{\textbf{Dataset}}} & \multicolumn{1}{c|}{\multirow{-2}{*}{\textbf{Model}}} &
\multicolumn{1}{c|}{\textbf{Specified}} &
\multicolumn{1}{c|}{\textbf{Violated}} &
\multicolumn{1}{c}{\textbf{Bugs}} \\ \midrule
\compas    & NN & 12 & 7  &   960.0 \\
           & DT & 12 & 6  &   294.8 \\
\credit    & NN & 10 & 6  &   295.2 \\
           & DT & 10 & 6  &   286.9 \\
\mnist     & NN &  1 & 1  &    22.4 \\ \hline
\speech    & NN &  1 & 1  &    14.2 \\ \hline
\hotel     & NN &  4 & 4  &  3288.0 \\ \hline
\lunar     & RL &  2 & 2  &  3459.0 \\
\end{tabular}
\label{tab:results-overview}
\end{table}

\textbf{Results.}
%
%
We specified $30$ properties across all datasets. Tab.~\ref{tab:results-overview} provides an overview of the number of specified properties, violated properties, and unique bugs per dataset and model. Our testing framework was able to find violations for all datasets, and in particular, for $24$ of these properties. Most property violations were exhibited through tens or hundreds of unique tests. This demonstrates that our framework is effective in detecting bugs even with as few as $5000$ tests per property; in contrast, fuzzers for software systems often generate millions of tests before uncovering a bug.

The average number of bugs per property varies significantly depending on the property, model, and dataset (see appendix for details). For instance, for \compas, the average number of bugs ranges from $0.5$ to $619.7$ when testing the NN classifier against each of the twelve different properties.

There are six properties that were not violated by any model trained on the \compas and \credit datasets. For four of these properties, we observed that the involved features almost never affect the outcome of our models, thereby trivially satisfying the properties. In the remaining cases, the training data itself seems to be sufficient in ensuring that the properties hold for the models.
%

\textbf{Feasibility study.}
Our results show that our framework is effective in detecting property violations. But are these violations actionable? A natural next step is to use them for repairing the model under test or incorporate them when training the model from scratch---much like adversarial training for robustness issues. While a comprehensive exploration and discussion of such options is beyond the scope of this work, we did perform a feasibility study to investigate whether the reported violations are indeed actionable.

\begin{table}[t]
\caption{Minimum-bug and maximum-reward policies generated with normal and guided training.}
\vspace{1em}
\centering
\begin{tabular}{S[table-format=2.0]|S[table-format=3.1]|S[table-format=2.0]|S[table-format=3.1]|S[table-format=2.0]|S[table-format=3.1]|S[table-format=2.0]|S[table-format=3.1]}
\rowcolor{light-gray!50}
\multicolumn{4}{c|}{\textbf{Minimum-Bug Policy}} & \multicolumn{4}{c}{\textbf{Maximum-Reward Policy}}\\
\rowcolor{light-gray!50}
\multicolumn{2}{c|}{Normal} & \multicolumn{2}{c|}{Guided} & \multicolumn{2}{c|}{Normal} & \multicolumn{2}{c}{Guided}\\
\rowcolor{light-gray!50}
\multicolumn{1}{c|}{Bugs} & \multicolumn{1}{c|}{Rew.} &
\multicolumn{1}{c|}{Bugs} & \multicolumn{1}{c|}{Rew.} &
\multicolumn{1}{c|}{Bugs} & \multicolumn{1}{c|}{Rew.} &
\multicolumn{1}{c|}{Bugs} & \multicolumn{1}{c}{Rew.}\\ \midrule
19 &    230.8  &    19  & \h{242.0} &    27  &    232.0  & \h{23} & \h{261.5} \\
12 &    155.5  &  \h{7} & \h{160.1} &    16  &    157.2  &  \h{8} & \h{197.0} \\
20 & \h{257.0} & \h{12} &    254.4  &    32  &    277.3  & \h{16} & \h{279.0} \\
19 &    170.2  &    19  &    170.2  &    29  &    175.0  & \h{27} & \h{184.5} \\
28 &  \h{83.7} & \h{16} &     62.9  & \h{29} &    137.2  &    34  & \h{167.7} \\
 8 & \h{237.4} &  \h{6} &    208.9  & \h{11} &    243.6  &    13  & \h{256.2} \\
21 &    224.8  & \h{12} & \h{254.7} &    29  &    240.8  & \h{21} & \h{264.1} \\
17 &     15.0  &  \h{7} & \h{220.2} &    24  &    181.7  & \h{12} & \h{221.6} \\
14 & \h{263.5} &  \h{9} &    209.0  & \h{14} & \h{263.5} &    23  &    242.4 \\
 9 &    128.1 &  \h{2} & \h{144.4}  &    16  &    158.7  &  \h{7} & \h{217.0} \\
\end{tabular}
\label{tab:training}
\end{table}

For this study, we selected \lunar due to its higher complexity. On a high level, we incorporated buggy game states, \ie, ones that resulted in property violations, during policy training. In particular, we adjusted the existing training algorithm (PPO~\cite{SchulmanWolski2017} implemented in the SB3 library~\cite{SB3}) to start episodes not only from random initial states, but also from buggy states. As training progresses, our guided-training algorithm gradually increases the probability of starting from buggy states. The intuition behind this choice is to focus more on "ironing out" bugs toward the end of the training, when the policy is already able to achieve decent rewards.

Under the hood, our guided-training algorithm tests the current policy at regular intervals (every $5$ rollouts in our experiments), essentially alternating between training and testing phases. Any bugs that are found during the latest testing phase are added to a pool of buggy states from which the algorithm selects initial states during subsequent training phases. Note that we prioritize most recently detected buggy states, but we also include older bugs to ensure the policy does not "forget" later on.

For our experiments, we trained $10$ policies with each training algorithm, \ie, normal and guided. Tab.~\ref{tab:training} summarizes the policies that were generated during these training runs---each row corresponds to a training run. In the four leftmost columns, we focus on policies with the fewest number of bugs. The first two columns show the number of bugs and reward for the minimum-bug policy generated during each of the normal-training runs. Note that, for policies with the same number of bugs during a run, we show the one with higher reward. Similarly, the third and fourth columns show the same data for guided training. In the four rightmost columns, we focus on policies with the highest reward. Again, for policies with the same reward, we show the one with fewer bugs.

Looking at the first and third columns, no normal-training run achieves fewer bugs than the corresponding guided-training run, and guided training results in fewer bugs in $8$ out of $10$ runs. Looking at the second and fourth columns, guided training does not result in significantly lower rewards for the minimum-bug policies; in $5$ out of $10$ runs, guided minimum-bug policies surpass, in terms of reward, the corresponding normal policies. In addition, when looking at the fourth and sixth columns, $4$ out of $10$ guided minimum-bug policies even surpass the normal maximum-reward policies. Similarly, when considering the maximum-reward policies, guided training results in higher rewards in $9$ out of $10$ runs; in $7$ runs, guided policies have fewer bugs; and $4$ guided maximum-reward policies have fewer bugs than the corresponding normal minimum-bug policies.

\begin{figure}[t]
    \centering
    \includegraphics[scale=0.5,clip]{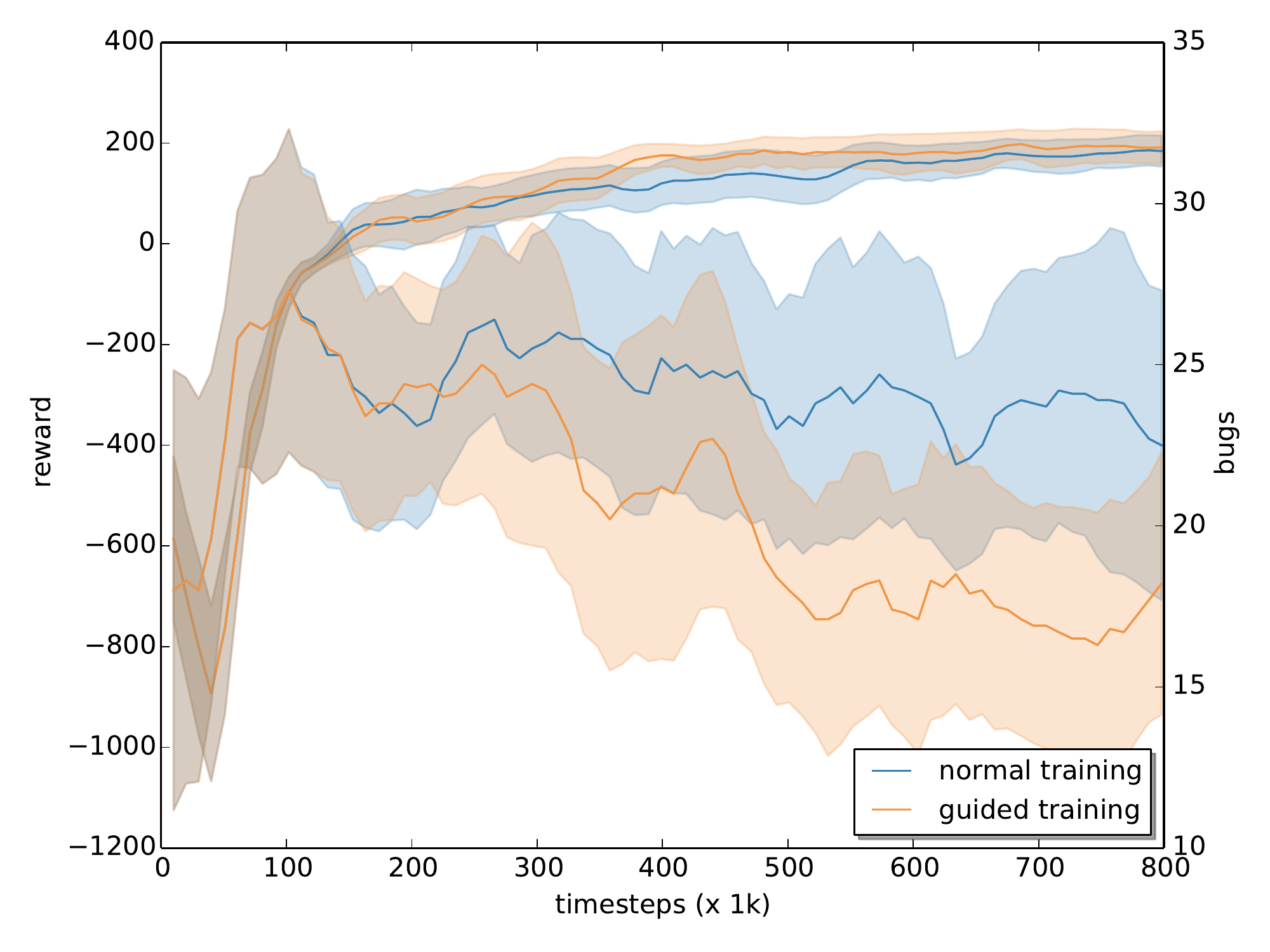}
    \caption{Increase in reward and decrease in number of bugs over time for normal and guided training.}
    \label{fig:retraining}
\end{figure}

\figref{fig:retraining} shows the increase in reward and decrease in number of bugs over time both for normal and guided training. The dark lines represent the mean values, and the lighter shaded areas denote the 90\% confidence interval. As expected, we observe that, for guided training, the number of bugs is consistently lower without compromising on the achieved reward.

\emph{Overall, our experiments show that property violations can be useful not only for assessing the quality of a model, but also for training better models.} The latter is a promising direction for future work.

\section{Conclusion and Outlook}\label{sec.conclusion}

We have presented the \lang language for specifying $k$-safety properties of ML models and an automated testing framework for detecting violations of such properties. \lang is the first high-level specification language for expressing general hyperproperties of models, subsuming more specific ones such as robustness and fairness. We have demonstrated the wide applicability of such properties through case studies from several domains and evaluated the effectiveness of our framework in detecting property violations.
Although users could manually write test cases or a test harness for each desired property, this would be tedious, repetitive, and easy to get wrong; it would also be difficult to update and extend properties if needed. In contrast, our \lang specifications are concise and enable users to think about properties on a higher level of abstraction.

There are several promising directions for future work. For the ML community, model repair and guided training might be the most interesting direction for building on \lang and our testing framework. One way to think about specifications is as a, possibly infinite, source of training examples. Our feasibility study has already provided some empirical evidence for how such examples can be incorporated in the training process. However, more work is needed, and adversarial-training techniques could be adapted to improve the effectiveness.

For the testing community, an interesting direction could be to explore more effective input-generation techniques, such as coverage-guided testing. This may reduce the testing time or increase the number of bugs that can be found within a given time budget. Such advances can be crucial for reducing the testing overhead when performing guided training.

For the formal-methods community, a natural next step is to build verification tools for certifying that a property holds \emph{for all inputs}. This could be particularly promising for models that are used in safety-critical domains, such as autonomous driving.

We believe that \lang can bring these communities together to facilitate the development of functionally correct ML models.

\bibliography{bibliography}
\iftoggle{longversion}{
\clearpage
\onecolumn
\appendix
\section{List of Appendices}\label{appendix.table-of-contents}

Below, we provide a brief description of each section in the appendix.
\begin{itemize}
    \item We provide additional grammar definitions in Appendix \ref{sect:appendix-grammar}.
    \item We include all \lang specifications for our case studies in Appendix \ref{sect:appendix-specs}.
    \item We present additional details of our experimental results (\eg, number of bugs for each specification) in Appendix \ref{sect:appendix-results}.
    \item We present additional details of our experimental setup (\eg, details on model training and hardware setup) in Appendix \ref{sect:appendix-addinfo}.
\end{itemize}

\section{Additional Grammar Definitions}
\label{sect:appendix-grammar}

Grammar definition for integer expressions in \lang:\\
\begin{lstlisting}[style=ebnf]
<int_expr>    ::= <int_literal>
                | <var_name>
                | "-"<int_expr>
                | <int_expr> "+" <int_expr>
                | <int_expr> "-" <int_expr>
                | <int_expr> "*" <int_expr>
                | <int_expr> "/" <int_expr>
\end{lstlisting}

Grammar definition for string expressions in \lang:\\
\begin{lstlisting}[style=ebnf]
<string_expr> ::= <string_literal>
                | <var_name>
\end{lstlisting}

We omit the basic scalar expressions \ebnf{<int_literal>} and \ebnf{<string_literal>}.

\section{Specifications}
\label{sect:appendix-specs}

Below, we provide a description of each specified property:
\begin{description}
\item[Felony Inc] If the number of committed felonies for a criminal increases, then their recidivism risk should not decrease.
\item[Felony Dec] If the number of committed felonies for a criminal decreases, then their recidivism risk should not increase.
\item[Misdmnr Inc] If the number of committed misdemeanors for a criminal increases, then their recidivism risk should not decrease.
\item[Misdmnr Dec] If the number of committed misdemeanors for a criminal decreases, then their recidivism risk should not increase.
\item[Priors Inc] If the number of priors for a criminal increases, then their recidivism risk should not decrease.
\item[Priors Dec] If the number of priors for a criminal decreases, then their recidivism risk should not increase.
\item[Others Inc] If the number of other crimes committed by a criminal increases, then their recidivism risk should not decrease.
\item[Others Dec] If the number of other crimes committed by a criminal decreases, then their recidivism risk should not increase.
\item[IsRecid Set] If a criminal becomes a recidivist, then their recidivism risk should not decrease.
\item[IsRecid Unset] If a criminal ceases to be a recidivist, then their recidivism risk should not increase.
\item[IsVRecid Set] If a criminal becomes a violent recidivist, then their recidivism risk should not decrease.
\item[IsVRecid Unset] If a criminal ceases to be a violent recidivist, then their recidivism risk should not increase.
\item[Crdt Amount Inc] If the credit amount requested by a person increases, then they should not be more likely to receive it.
\item[Crdt Amount Dec] If the credit amount requested by a person decreases, then they should not be less likely to receive it.
\item[Crdt Hist Inc] If a person's credit history worsens, then they should not be more likely to receive credit.
\item[Crdt Hist Dec] If a person's credit history improves, then they should not be less likely to receive credit.
\item[Empl Since Inc] If a person's employment years increase, then they should not be less likely to receive credit.
\item[Empl Since Dec] If a person's employment years decrease, then they should not be more likely to receive credit.
\item[Install Rate Inc] If a person's installment rate (as a percentage of their disposable income) increases, then they should not be more likely to receive credit.
\item[Install Rate Dec] If a person's installment rate (as a percentage of their disposable income) decreases, then they should not be less likely to receive credit.
\item[Job Inc] If a person is promoted, then they should not be less likely to receive credit.
\item[Job Dec] If a person is demoted, then they should not be more likely to receive credit.
\item[Blur] If a blurred image is correctly classified, then its unblurred version should also be correctly classified.
\item[WNoise] If a speech command with white noise is correctly classified, then its non-noisy version should also be correctly classified.
\item[Pos-1] Deleting the positive comments of a hotel review should not make it more positive.
\item[Pos-2] If more positive comments are added to a hotel review, it should not become more negative.
\item[Neg-1] Deleting the negative comments of a hotel review should not make it more negative.
\item[Neg-2] If more negative comments are added to a hotel review, it should not become more positive.
\item[Relax] If the lander lands successfully, then decreasing the surface height (thus giving the lander more time to land) should also result in landing successfully.
\item[Unrelax] If the lander fails to land, then increasing the surface height (thus giving the lander less time to land) should also result in failing to land.
\end{description}

\section{Additional Details of our Experimental Results}
\label{sect:appendix-results}

In this section, we provide more detailed results on the number of unique bugs for each individual specification.

The results for \compas are shown in Tab.~\ref{tab:perspec-compas}. Column 2 shows the average number of unique bugs for the NN model and for each of the 12 specifications. Column 3 shows the same data for the DT model.
For properties "IsRecid Set" and "IsRecid Unset", we observed that the involved feature "IsRecid" almost never affects the outcome of our models, thereby trivially satisfying the properties.

Tab.~\ref{tab:perspec-credit} shows similar results for \credit.
For properties "Install Rate Inc" and "Install Rate Dec", we observed that the involved feature "Installment Rate" almost never affects the outcome of our models, thereby trivially satisfying the properties.

Tab.~\ref{tab:perspec-complex} shows similar results for the remaining benchmarks.

\begin{table}[ht]
\centering
\caption{Average number of unique bugs for each \compas specification.}
\vspace{0.5em}
\label{tab:perspec-compas}
\begin{tabular}{l|S[table-format=3.1]|S[table-format=3.1]}
\rowcolor{light-gray!50}
& \multicolumn{2}{c}{\textbf{Unique Bugs}}  \\
\rowcolor{light-gray!50}
\multicolumn{1}{c|}{\multirow{-2}{*}{\textbf{Specification}}} & \multicolumn{1}{c|}{NN} & \multicolumn{1}{c}{DT} \\ \midrule
Felony Inc     &  42.9 &  3.5 \\ 
Felony Dec     &   0.0 &  0.0 \\
Misdmnr Inc   & 619.7 &  0.0 \\
Misdmnr Dec   &   4.0 &  0.0 \\
Priors Inc     &   0.5 & 90.0 \\
Priors Dec     &   0.0 & 97.2 \\
Others Inc     & 289.0 & 91.1 \\
Others Dec     &   3.0 &  8.0 \\
IsRecid Set    &   0.0 &  0.0 \\
IsRecid Unset &   0.0 &  0.0 \\
IsVRecid Set   &   0.9 &  5.0 \\
IsVRecid Unset &   0.0 &  0.0 \\
\end{tabular}
\end{table}

\begin{table}[ht]
\centering
\caption{Average number of unique bugs for each \credit specification.}
\vspace{0.5em}
\label{tab:perspec-credit}
\begin{tabular}{l|S[table-format=3.1]|S[table-format=3.1]}
\rowcolor{light-gray!50}
& \multicolumn{2}{c}{\textbf{Unique Bugs}}  \\
\rowcolor{light-gray!50}
\multicolumn{1}{c|}{\multirow{-2}{*}{\textbf{Specification}}} & \multicolumn{1}{c|}{NN} & \multicolumn{1}{c}{DT} \\ \midrule
Crdt Amount Inc      & 0.0   & 122.3 \\
Crdt Amount Dec      & 0.0   &  75.4 \\
Crdt Hist Inc      & 78.1  &   8.0 \\
Crdt Hist Dec      & 122.8 &  31.2 \\
Empl Since Inc     & 13.5  &   9.1 \\
Empl Since Dec     & 30.9  &  40.9 \\
Install Rate Inc    & 0.0   &   0.0  \\
Install Rate Dec    & 0.0   &   0.0  \\
Job Inc            & 47.9  &   0.0  \\
Job Dec            & 2.0   &   0.0  \\
\end{tabular}
\end{table}

\begin{table}[hb]
\centering
\caption{Average number of unique bugs for all \mnist, \speech, \hotel, and \lunar specifications.}
\vspace{0.5em}
\label{tab:perspec-complex}
\newcolumntype{g}{>{\columncolor{light-gray!20}}r}
\newcolumntype{h}{>{\columncolor{light-gray!20}}c}
\begin{tabular}{g|l|S[table-format=4.1]}
\rowcolor{light-gray!50}
\multicolumn{1}{c|}{\textbf{Benchmark}} & \multicolumn{1}{c|}{\textbf{Specification}} & \multicolumn{1}{c}{\textbf{Unique Bugs}} \\ \midrule
\mnist  &  Blur    &   22.4 \\ \hline
\speech &  WNoise  &   14.2 \\ \hline
\hotel   &  Pos-1   &  861.1 \\
  &  Pos-2   &  876.1 \\
  &  Neg-1   &  756.2 \\
  &  Neg-2   &  794.6 \\ \hline
\lunar  &  Relax   &  124.5 \\
  &  Unrelax & 3334.5 \\
\end{tabular}
\end{table}

\section{Additional Details of our Experimental Setup}
\label{sect:appendix-addinfo}

\subsection{Training Setup}
For the \compas dataset, we trained a fully connected neural network and a decision tree classifier. The neural network is composed of 3 hidden layers of size 12, 9, and 9. We use the RMSprop algorithm for optimization. For decision-tree training, we set the $\mathit{max\_depth}$ parameter to 8. For training, we shuffle the data and use 67\% of it.

For the \credit dataset, we trained a fully connected neural network and a decision tree classifier. The neural network is composed of 1 hidden layer of size 10, and we use the Adam optimizer. In decision-tree training, we set the $\mathit{max\_depth}$ parameter to 6. For training, we shuffle the data and use 67\% of it.

For the \mnist dataset, we trained a fully connected neural network consisting of 3 hidden layers (each with 30 neurons), and we use the Adam optimizer. We use the regular training set for training.

For the \speech dataset, we apply a number of pre-processing steps and infer a spectrogram image for each audio file. We use 80\%  of the spectogram inputs for training a convolutional neural network consisting of 2 convolutional layers with kernels (32x32x3) and (64x64x3), and a fully connected layer of size 128. We use dropout for regularization and Adam for optimization.

The \hotel dataset consists of over 515k reviews, and only ca. 85k of them are scored above 6 (out of 10)---labeled as \emph{positive} in our evaluation. We sample the same number of inputs from the ones that are labeled as \emph{negative} to form a new dataset consisting of around 170k inputs. We use 90\% of them as training set. We use the USE model from Tensorflow Hub\footnote{https://tfhub.dev/google/universal-sentence-encoder-multilingual-large/3}. The USE-encoded reviews are used to train a fully connected neural network with 2 hidden layers (256 and 128 neurons, respectively). We use dropout for regularization and Adam for optimization.

For the \lunar dataset, we use the default PPO implementation in the SB3 library for training the agent.


We use ReLU activation functions in all neural networks.

We use Tensorflow v2.7 and the scikit-learn v1.0.2 framework for training neural networks and decision trees, respectively.

\subsection{Hardware Setup}

We use a cluster with a Quadro RTX 8000 GPU and an Intel(R) Xeon(R) Gold 6248R CPU @ 3.00GHz for training models and running tests. Running 5k tests takes a few seconds for decision trees. It takes longer for neural networks, ranging from 5 to 20 minutes depending on the specification and the dataset. For \lunar, it takes up to 4 hours.

The total amount of compute for all experiments is ca. 1 day on the above cluster.

}
{
}


\end{document}